\documentclass[runningheads]{llncs}
\usepackage[T1]{fontenc}
\usepackage{graphicx}
\usepackage{booktabs}
\usepackage[misc]{ifsym}
\newcommand{\corr}{(\Letter)}
\usepackage{mwe}

\usepackage{amsmath}
\usepackage{rotating}
\usepackage{algorithm}
\usepackage{algorithmic}
\usepackage{amsfonts}
\usepackage{tikz}
\usepackage{booktabs}
\usepackage{multirow}
\usepackage{array, makecell}
\usepackage{color, colortbl}
\usepackage[table]{xcolor}
\usepackage{multirow}
\usepackage{rotating}
\definecolor{LightGrey}{rgb}{0.7,0.7,0.7}
\usepackage{tabularray}  
\usepackage{xcolor}
\usepackage{amsmath}
\DeclareMathOperator*{\argmin}{arg min}
\DeclareMathOperator*{\W}{\mathcal{W}}
\DeclareMathOperator*{\Loss}{\mathcal{L}}

\usepackage[hidelinks]{hyperref}
\usepackage{newfloat}
\usepackage{listings}
\floatstyle{ruled}
\newfloat{listing}{tb}{lst}{}
\begin{document}

\renewcommand{\thefootnote}{}
\footnotetext{A. Prutianova, A. Bogdanova~--- Equal contributions.}

\title{Trainable Vector Quantization for Large Language Models}
\toctitle{Trainable Vector Quantization for Large Language Models}
\titlerunning{Trainable Vector Quantization for Large Language Models}
% If the full title of your paper is short enough to also fit in the running head, you can omit the abbreviated paper title here. You can check as follows: if you comment out the \titlerunning line, something will appear in the header of all odd-numbered pages of your PDF from page 3 onward. This something is either the full title (in which case all is well), or the error message "Title Suppressed Due to Excessive Length". If this error message appears, you're going to want to provide an abbreviated title within the \titlerunning command, because if you won't do it, Springer will do it for you.

%N.B.: Author information (both in the \author{} and \authorrunning{} command) should only be present in the Camera-Ready Version of your paper. The version that you initially submit for review, ought to be double-blind. So, when initially submitting your paper, use:
%\author{Author information scrubbed for double-blind reviewing}
\author{
    Anastasiia Prutianova \and
    Alina Bogdanova \and
    Mikhail Shutov \and
    Dmitrii Poimanov \and
    Anatoly Gorkov \and
    Ilia Zharikov\corr
}
\tocauthor{Anastasiia Prutianova, Alina Bogdanova, Mikhail Shutov, Dmitrii Poimanov, Anatoly Gorkov, Ilia Zharikov}
\authorrunning{A. Prutianova, A. Bogdanova et al.}
% You may leave out the orcidID information, if you want to.
% Use \corr to indicate the corresponding author. Note the spacing around the \corr command. Only one author can be the corresponding author.

%N.B.: comment out the \authorrunning{} command for the double-blind version of your paper submitted for review. Later, if your paper is accepted, use the command for the Camera-Ready Version.
% If there are two authors, write 'A.L. Benjamin and C.C. Broadus Jr.'
% If there are more than two authors, '[...] et al.' is used.

\institute{Huawei Technologies Co. Ltd. \\
\email{ilya.zharikov@huawei.com}}

\maketitle              % typeset the header of the contribution

\begin{abstract}
Vector Quantization (VQ) is a rapidly evolving technique for compressing Large Language Model (LLM) weights, reducing memory usage, and speeding up inference. However, extreme compression deteriorates model performance, which is challenging to recover due to the discrete nature of VQ parameters. Existing methods employ training-free approaches that are limited to layer-wise reconstruction loss minimization. To address this limitation, we propose TVQ, a novel method that leverages straight-through estimation to enable gradient descent-based optimization of vector-quantized LLMs. By reformulating VQ LLM optimization with trainable latent weights, TVQ enables arbitrary objective minimization. We constitute our approach by applying TVQ in a block-wise knowledge distillation paradigm, which balances quality and training efficiency, and propose several practical solutions to accelerate training. In experiments on modern LLMs, we demonstrate that TVQ is compatible with modern VQ schemes. As a result, TVQ achieves state-of-the-art quality in low-bit post-training quantization settings with a modest training budget and straightforward PyTorch implementation. For instance, TVQ achieves near-lossless 2-bit quantization for LLaMA2-70B with only 1\% accuracy drop on popular Question-Answering benchmarks.
\end{abstract}

% Fig.\@ \ref{fig1}
\section{Introduction}
Recent advancements in deep learning have led to significant progress in Natural Language Processing, particularly in text generation~\cite{achiam2023gpt,dubey2024llama3,guo2025deepseek}.  A significant factor driving this progress is the rapid increase in neural network size, which leads to enhanced performance but also raises computational costs and memory consumption, fueling demand for efficient language models. 

Large Language Model (LLM) deployment is primarily memory-bound; for instance, the \mbox{Llama3-70B~\cite{dubey2024llama3}} parameters require 140GB of memory, exceeding the capacity of modern Graphics Processing Units (GPU). 
As weight-only quantization allows for compressing the model several-fold, it has become crucial for
efficient memory-constrained inference~\cite{frantar2023gptq,xiao2023smoothquant,dettmers2023qlora}. In particular, Post-Training Quantization (PTQ)~\cite{adaround2020,adaquant2021,lin2024awq,chen2024efficientqat} was widely adopted since it compresses pretrained weights without full model retraining, adding only negligible computational overhead for calibration. 

Recently, Vector Quantization (VQ) has gained traction as it demonstrated significant memory reduction advantages over traditional scalar quantization~\cite{gptvq,DBLP:conf/icml/EgiazarianPKFBA24}. Unlike scalar quantization~\cite{pact2018,LSQ,158bit2024}, which encodes values of the weight matrix with a fixed set of integers, VQ represents the weight matrix using a fixed set of vectors, known as a codebook. 
Specifically, in VQ, the values of the pretrained weight matrix are grouped into fixed-length vectors, and each group is replaced with a codebook vector.
An integer index matrix is then used to indicate which codebook vector is assigned to each group.

Since extreme compression inevitably introduces a performance gap relative to the original model, it is necessary to optimize the quantization parameters to recover model quality. The key challenge is that the majority of VQ parameters are concentrated in integer index matrices, resulting in a high-dimensional and discrete search space. Most existing VQ optimization methods~\cite{DBLP:conf/icml/EgiazarianPKFBA24,liu_vptq_2024,tseng2024quip,tseng2024qtip} simplify the vector assignment optimization problem by considering individual model layers independently. These methods aim to minimize the quantization error of linear layer outputs, specifically $\displaystyle{\left\Vert(W^q - W)X\right\Vert_2}$, where $\displaystyle{W^q}$ and  $\displaystyle{W}$ are compressed and original weight matrices, $\displaystyle{X}$ is a matrix of input activations. They rely on single-layer linearity and $\displaystyle{L_2}$ layer output loss as the primary objective. As a result, network layers have a limited ability to adapt to the quantization of other layers and cannot capture how local quantization errors propagate through subsequent layers.

We address this challenge by introducing Trainable Vector Quantization (TVQ), a method that enables simultaneous optimization of multiple VQ parameters with respect to arbitrary loss functions. To achieve this, we reformulate the index matrix optimization problem by introducing trainable continuous latent weights and leveraging the Straight-Through Estimator (STE)~\cite{bengio2013estimating} to propagate gradients through discrete index assignments from quantized to latent weights.

Although TVQ allows simultaneous optimization of all parameters, full model training is resource-intensive. Instead, we chose to use TVQ within the block-wise Knowledge Distillation (KD) paradigm~\cite{li2021brecq,adaquant2021,wei2022qdrop,omniquant2024,chen2024efficientqat}, which is known to be state-of-the-art in scalar PTQ in both computer vision and language domains. However, it remains unexplored in VQ LLMs due to the limitations of prior VQ optimization methods. Block-wise KD strikes a good balance between training speed and quality and requires only a small amount of calibration data. As a result, our block-wise TVQ achieves computational efficiency comparable to the layer-wise methods, while delivering significant quality improvements. Furthermore, TVQ outperforms the model-wise fine-tuning approach PV-tuning~\cite{malinovskii2024pv} in both speed and quality. Additionally, we suggest a straightforward implementation of TVQ with popular training frameworks such as PyTorch~\cite{paszke2019pytorch}.

TVQ is suitable for modern advanced VQ schemes~\cite{tseng2024quip,tseng2024qtip} introduced in previous works. We experimentally demonstrate that TVQ improves the model performance via better optimization while keeping the architecture of VQ LLM unchanged. As a result, models optimized with TVQ remain compatible with existing software tools for inference acceleration.

In summary, our contributions are the following:

\begin{itemize}
\item  We introduce a novel VQ LLM optimization method TVQ, that enables gradient descent-based optimization of vector-quantized weights with respect to an arbitrary loss function. We propose to reformulate index assignment optimization via trainable continuous latent weights and enable backpropagation through the discrete assignment operation with a STE. We experimentally demonstrate that TVQ is applicable to existing advanced VQ schemes, computationally efficient, and easy to implement.

\item In extensive experiments on modern LLMs, such as \textsc{Llama 2}~\cite{touvron2023llama2}, \textsc{Mistral}~\cite{jiang2023mistral}, and \textsc{Llama 3}~\cite{dubey2024llama3}, we show that TVQ outperforms existing optimization techniques. As a result, it achieves state-of-the-art performance in low-bit PTQ.  Notably, for 2-bit Llama3-70B, TVQ improves upon the current best result by 1.9\% of accuracy across popular Question-Answering benchmarks, substantially reducing the performance gap between the original and quantized models from 4.5\% to 2.6\%. In addition, we provide detailed ablation studies on the impact of the proposed method's components.
\end{itemize}

\section{Related Work}
\subsection{Vector-Quantized LLM Optimization}

This section focuses on the techniques employed in related works, specifically for optimizing vector-quantized LLMs. Existing approaches follow a two-stage pipeline: layer-wise optimization that minimizes reconstruction loss, followed by fine-tuning with respect to global objectives.

\textbf{Stage-1: Layer-wise Optimization.} Prior methods start by optimizing the reconstruction error $\displaystyle{\left\Vert(W^q - W)X\right\Vert_2}$, where they jointly determine vector assignments and codebooks for a single linear layer. AQLM~\cite{DBLP:conf/icml/EgiazarianPKFBA24} employs an EM-algorithm that alternates between beam search for index matrix (E-step) and gradient descent for codebook optimization (M-step), with codebook initialization via standard k-means clustering. The VPTQ~\cite{liu_vptq_2024} achieves comparable quality with reduced training time via enhanced Hessian-weighted k-means codebook initialization and GPTQ-like~\cite{frantar2023gptq} adaptive rounding approach for vector assignment. However, both AQLM and VPTQ are incompatible with the fixed hardware-efficient codebooks employed in QuIP\#~\cite{tseng2024quip} and QTIP~\cite{tseng2024qtip}. These immutable codebooks prevent M-step updates in AQLM and weighted k-means initialization in VPTQ. To optimize vector assignments with fixed codebooks, QuIP\# and QTIP leverage the BlockLDLQ algorithm, which employs Hessian-aware adaptive rounding based on the LDL decomposition.
However, both VPTQ and BlockLDLQ estimate the Hessian solely from layer inputs and therefore cannot account for error propagation through subsequent layers. 

\textbf{Stage-2: End-to-end Fine-tuning.} To mitigate the accumulated quantization errors from the layer-wise stage, existing works fine-tune continuous parameters (codebooks, layer norms, LM head) for a global objective such as cross-entropy or KD losses, while keeping vector assignments fixed. PV-tuning~\cite{malinovskii2024pv}, designed to complement Stage-1 algorithms, is the only method that considers model-wise vector assignment optimization, using gradient magnitude to select a small fraction of vectors for updates. However, PV-tuning requires significantly more compute than competing methods and does not naturally generalize to the current state-of-the-art VQ scheme, suggested in QTIP, where index matrices are bitshift trellis encoded and changing individual indices would break their codependence.

In contrast, TVQ is applicable to all existing VQ schemes. Unlike existing Stage‑1 approaches, which operate strictly in a layer-wise manner and cannot be directly extended to multi-layer optimization, TVQ captures inter-layer dependencies. For Stage-2 optimization, we apply continuous parameter fine-tuning as in most prior works.

\subsection{Straight-Through Estimation}
STE is a method that propagates gradients through non-differentiable or zero-gradient functions, allowing for gradient-based optimization. It was used in many areas where such functions naturally appear: in neural architecture search~\cite{dong2019searching}, pruning~\cite{vanderschueren2023straight}, scalar quantization~\cite{LSQ}. It was widely used in computer vision and speech processing for learning codebooks~\cite{DBLP:conf/nips/OordVK17,Stock2020And,NEURIPS2024_9a24e284}. However, STE has not previously been applied to the optimization of VQ LLMs; we address this gap by introducing TVQ.

\section{Method}
\subsection{Vector Quantization Background}\label{subsec: Vector Quantization Background}
Let the weight matrix of a linear layer be denoted by $\displaystyle{W \in \mathbb{R}^{d_{\text{out}}\times d_{\text{in}}}}$. For vector quantization this matrix is partitioned into $\displaystyle{d}$-dimensional vectors $\displaystyle{w_{ij} \in \mathbb{R}^d}$, forming a set of $\displaystyle{d_{\text{out}} \cdot d_{\text{in}} / d}$ vectors:

\begin{equation}
    W = \Bigl\{w_{ij} \Bigr\}_{i=1, j=1}^{d_{\text{out}}, \frac{d_{\text{in}}}{d}}
\end{equation}

For compression, each vector $\displaystyle{w_{ij}}$ is replaced with one of $\displaystyle{n_C}$ representative vectors stored in the codebook matrix $\displaystyle{C~\in~\mathbb{R}^{n_C \times d}}$. 
Index matrix $I$ specifies indices of the codebook vectors assigned to each vector $\displaystyle{w_{ij}}$. Since matrices $\displaystyle{I}$ and $\displaystyle{C}$ together contain significantly fewer parameters than $\displaystyle{W}$ does, they form a compressed representation of $\displaystyle{W}$. During inference, an uncompressed representation of $\displaystyle{W}$, denoted by $\displaystyle{W^q}$, is constructed from $\displaystyle{I}$ and $\displaystyle{C}$ as follows: 

\begin{equation}
    \label{eq:quantized_weight}
    W^q(I, C) = \Bigl\{w^q_{ij} \Bigm| w^q_{ij} = C_{I_{ij}} \Bigr\}_{i=1, j=1}^{d_{out}, \frac{d_{in}}{d}}
\end{equation}

Notably, $\displaystyle{W^q}$ is not necessarily the closest approximation of the original weight $\displaystyle{W}$, as the compressed representation $\displaystyle{\left(C, I\right)}$ is optimized to improve overall model performance. 

Additionally, some works apply pre-/post-processing functions to $\displaystyle{W}$ and $\displaystyle{W^q}$, respectively. Since pre-/post-processing and codebook optimization do not influence the logic behind our proposed index optimization method, TVQ, we omit them from the discussion in the next sections for simplicity.

\subsection{VQ schemes}\label{subsec: vq schemes}

To introduce a general TVQ framework, we begin by formally describing existing VQ schemes from the optimization perspective.

In plain VQ, the codebook can be any matrix $\displaystyle{C \in \mathbb{R}^{n_C \times d}}$, and the index matrix can take any values $\displaystyle{I \in \{0, \dots, n_C - 1\}^{d_{\text{out}} \times \frac{d_{\text{in}}}{d}}}$. We refer to this unconstrained scheme as BasicVQ. Recent work has proposed more advanced VQ schemes to enhance quality-acceleration trade-offs. From the optimization perspective, these methods restrict the search spaces for the codebook and index matrices to $\displaystyle{\mathcal{C} \subseteq \mathbb{R}^{n_C \times d}}$ and $\displaystyle{\mathcal{I} \subseteq \{0, \dots, n_C - 1\}^{d_{\text{out}} \times \frac{d_{\text{in}}}{d}}}$, respectively. For instance, the codebook matrix $\displaystyle{C}$ may be fixed and predefined to fit into the L1 GPU cache~\cite{tseng2024quip}, meaning that $\displaystyle{\mathcal{C}}$ contains only one element. Alternatively, $\displaystyle{\mathcal{I}}$ may consist of index matrices whose entries are codependent for advanced bit-packing techniques~\cite{tseng2024qtip} that improve storage efficiency. For clarity, we provide a summary of the existing VQ schemes in Appendix~\ref{subsec: appendix vq schemes}.

% \begin{figure*}[t]
% \centering
% \includegraphics[width=0.85\textwidth]{scheme_color.pdf} % 
% \caption{This figure illustrates the proposed TVQ method, applied to a single linear layer with the weight matrix $\displaystyle{W}$. During optimization, these matrices are updated: the indices matrix $\displaystyle{I}$, the weight matrix $\displaystyle{W}$, and (optionally) the codebook $\displaystyle{C}$. TVQ involves three main operations: $\displaystyle{I}$ is calculated according to~\eqref{eq:I_from_W}; The dequantized weight matrix $\displaystyle{W^q}$ is gathered from $\displaystyle{C}$ and $\displaystyle{I}$ according to~\eqref{eq:quantized_weight} and used to calculate the loss function; The gradients of the loss are propagated from $\displaystyle{W^q}$ to $\displaystyle{W}$ and $\displaystyle{C}$, $\displaystyle{W}$ (and optionally $\displaystyle{C}$) is updated with gradient descent. 
% After training, only the compressed representation consisting of $\displaystyle{C}$ and $\displaystyle{I}$ is stored and used for inference instead of the original weight matrix.}
% \label{fig: TVQ algorithm}
% \end{figure*}

\subsection{Trainable Vector Quantization}\label{subsec: Trainable Vector Quantization}

In this section, we present the general Trainable Vector Quantization (TVQ) algorithm, which enables joint optimization of VQ parameters across multiple layers with respect to an arbitrary loss function, and is compatible with existing LLM VQ schemes.

The compressed representation of a weight matrix consists of a codebook matrix $\displaystyle{C}$ and an index matrix $\displaystyle{I}$, with the majority of parameters concentrated in the index matrix (see Appendix~\ref{subsec: Bits/weight details}). 
Therefore, optimizing $\displaystyle{I}$ is crucial for recovering the model's quality. A key challenge is that the index matrix $\displaystyle{I}$ consists of integer values, resulting in a high-dimensional, combinatorial discrete search space and making direct optimization via standard gradient descent inapplicable. 
Nevertheless, leveraging gradient descent is highly desirable, as it provides a general mechanism to optimize models with respect to an arbitrary differentiable loss function $\mathcal{L}$.

To address this issue, we propose to reparameterize the index matrix $\displaystyle{I}$ as a function of the codebook $\displaystyle{C}$ and a continuous latent weight matrix $\displaystyle{W}$ as follows:

\begin{equation}
\label{eq:I_from_W}
    I\left(W, C\right) = \argmin_{I \in \mathcal{I} } \left\|W - W^q(I, C) \right\|_2  
\end{equation}

This function determines the indices that provide the closest approximation of the latent weight
$\displaystyle{W}$ with codebook $\displaystyle{C}$ under the VQ scheme constraints.

\begin{figure*}[t]
\centering
\includegraphics[width=0.85\textwidth]{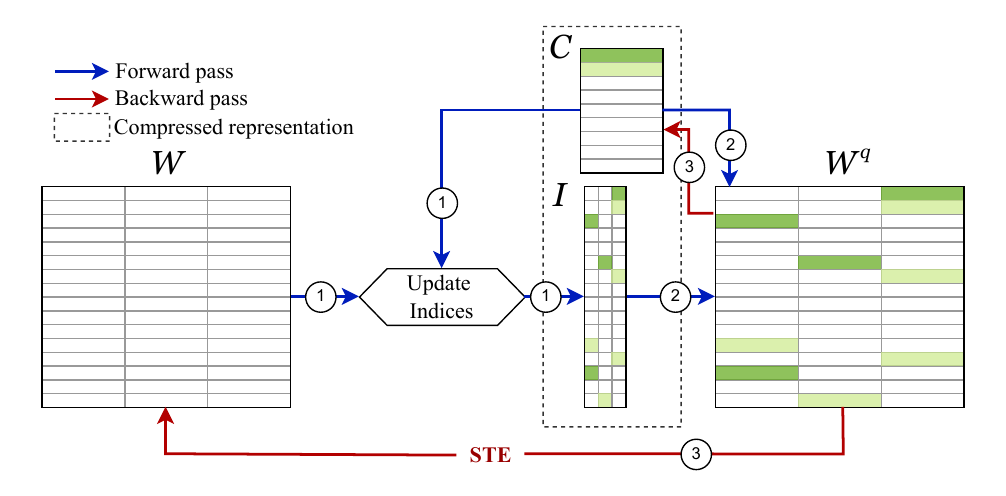} % 
\caption{This figure illustrates the proposed TVQ method, applied to a single linear layer with the weight matrix $\displaystyle{W}$. During optimization, these matrices are updated: the indices matrix $\displaystyle{I}$, the weight matrix $\displaystyle{W}$, and (optionally) the codebook $\displaystyle{C}$. TVQ involves three main operations: $\displaystyle{I}$ is calculated according to~\eqref{eq:I_from_W}; The dequantized weight matrix $\displaystyle{W^q}$ is gathered from $\displaystyle{C}$ and $\displaystyle{I}$ according to~\eqref{eq:quantized_weight} and used to calculate the loss function; The gradients of the loss are propagated from $\displaystyle{W^q}$ to $\displaystyle{W}$ and $\displaystyle{C}$, $\displaystyle{W}$ (and optionally $\displaystyle{C}$) is updated with gradient descent. 
After training, only the compressed representation consisting of $\displaystyle{C}$ and $\displaystyle{I}$ is stored and used for inference instead of the original weight matrix.}
\label{fig: TVQ algorithm}
\end{figure*}

This reparameterization transforms optimization over the discrete index matrix $\displaystyle{I}$ into optimization over the continuous latent weight $\displaystyle{W}$.
As a result, $\displaystyle{W^q}$ becomes a differentiable function of continuous parameters $\displaystyle{W}$ and $\displaystyle{C}$, thereby facilitating gradient-based optimization. The gradient of $\mathcal{L}$ with respect to $\displaystyle{W}$ is defined as follows:

\begin{equation}
    \label{eq:gradient_w}
    \frac{\partial\!\Loss}{\partial W} = \frac{\partial\!\Loss}{\partial W^q} \cdot \frac{\partial W^q}{\partial W}
\end{equation}

However, $\displaystyle{\partial W^q / \partial W}$ is equal to $0$ almost everywhere, since~\eqref{eq:I_from_W} is a piecewise constant function of $\displaystyle{W}$. According to~\eqref{eq:gradient_w}, this yields $\displaystyle{\partial \mathcal{L} / \partial W = 0}$, which results in zero gradient descent updates for $\displaystyle{W}$ and prevents learning. 

We propose to use STE to propagate gradients directly from $\displaystyle{W^q}$ to $\displaystyle{W}$. We replace a.e. zero gradient $\displaystyle{\partial W^q / \partial W}$ with the gradient of the identity function. Using this approximation in~\eqref{eq:gradient_w} we obtain the final gradient estimation, which we use for gradient descent updates of $\displaystyle{W}$: 

\begin{equation}
    \label{eq:gradient_w_ste}
    \frac{\partial\!\Loss}{\partial W} = \frac{\partial\!\Loss}{\partial W^q} \cdot \frac{\partial W^q}{\partial \W} \mathrel{\stackrel{\rm STE}=} \frac{\partial\!\Loss}{\partial W^q}
\end{equation}

The proposed idea has a very simple implementation in PyTorch or TensorFlow frameworks (see Appendix~\ref{subsec: TVQ implementation} for the suggested pseudocode). The general scheme of TVQ is presented in Fig.~\ref{fig: TVQ algorithm}.

\subsection{Training pipeline}\label{subsec: training pipeline}
\label{subsec:training_pipeline}
Our optimization pipeline consists of two stages. In the first stage, we apply TVQ to perform block-wise optimization of vector assignments. In the second stage, we conduct end-to-end model finetuning, adjusting only a small subset of parameters.

\subsubsection{Stage-1: Block-wise PTQ.}
A modern Large Language Model $\displaystyle{\mathcal{M}}$ is typically composed of $\displaystyle{K}$ sequential Transformer blocks:

\begin{equation}
\mathcal{M} = B_K \circ B_{K-1} \circ \dots \circ B_1 ,
\end{equation}
where $\displaystyle{B_k(h_{k-1}, \theta_k)}$ denotes the $\displaystyle{k}$-th Transformer block, consisting of an attention module, a following multi-layer perceptron, and residual connections. The input activations to the $\displaystyle{k}$-th block and its parameters are denoted as $\displaystyle{h_{k-1}}$ and $\displaystyle{\theta_k}$, respectively. 
Similarly, a quantized model consisting of quantized blocks $\displaystyle{B^q_i}$ is defined in a similar way:

\begin{equation}
    \mathcal{M}^q = B^{q}_{K} \circ B^{q}_{K-1} \circ \dots \circ B^q_{1}.
\end{equation}

We optimize the quantized model blocks sequentially by minimizing the distance between their output activations and corresponding activations of the original model. Index matrices are trained using TVQ to reduce the reconstruction loss on the outputs,

\begin{equation}
    \min_{\theta^q_k} {\left\Vert B^q_{k}\left(h^q_{k-1}, \theta^q_k\right) - B_k\left(h_{k-1}, \theta_k\right) \right\Vert_p},
\end{equation}
where $\displaystyle{h_{k-1} = B_{k-1} \circ \dots \circ B_1 (X)}$ is input activations to block $\displaystyle{k}$ computed by the original model, $\displaystyle{h^q_{k-1} = B^q_{k-1} \circ \dots \circ B^q_{1} (X)}$ is input activations to block $\displaystyle{k}$ computed by the quantized model, $\displaystyle{X}$ refers to the input training data.

\subsubsection{Stage-2: Parameter efficient End-to-End Fine-Tuning.}

In this stage, we optimize the quantized model $\displaystyle{\mathcal{M}^q}$ continious parameters with gradient decent. We use Kullback-Leibler (KL) divergence as a loss function,

\begin{equation}
    \label{eq:e2e}
    \min_{\theta_s^q}\mathrm{KL} \left( \mathcal{M}^q\left(X, \theta^q\right), \mathcal{M}\left(X, \theta\right) \right),
\end{equation}
where $\displaystyle{\theta_s^q \subset \theta^q}$ is a selected subset of the VQ model’s continuous parameters, such as layer normalization weights, codebooks, and the language modeling head. We optimize parameter subsets similar to those used in prior works, which vary depending on the specific quantization scheme employed.

\subsection{Implementation details}\label{subsec: implementation details}

TVQ requires recomputing the index matrices at each optimization step based on the updated latent weights, as defined in Equation~\eqref{eq:I_from_W}.  
The index computation algorithm depends on the optimization constraints imposed by the chosen VQ scheme and can be computationally expensive (see  Appendix~\ref{subsec: Index updates computational costs} for details about computational overhead for different VQ schemes). To address this, we suggest two complementary acceleration strategies:
\begin{itemize}
    \item Scheduled Index Updates. In some experiments, we update the index matrices once in $\displaystyle{\mathbf{S}_{\text{upd}}}$ optimization steps (rather than at each step), since it has a negligible impact on the final model quality, as we confirm in Section~\ref{subsec: ablation scheduled updates}.
    Intuitively, these scheduled updates are similar to increasing the effective batch size via gradient accumulation.

    \item Approximate Nearest-Neighbor Search. For TVQ with the basic VQ scheme defined in Section~\ref{subsec: Vector Quantization Background} (referred to as TVQ+BasicVQ in the experimental results), we update index matrices via Approximate Nearest-Neighbor search. We utilize an approximate search~\cite{kolahdouzan2004voronoi} based on Voronoi cells partitioning, implemented in the FAISS library~\cite{faiss}. As we show in Appendix~\ref{sec: ANN ablation}, it has high precision and does not affect the final model quality while speeding up index matrix updates.
\end{itemize}
For further details on training acceleration see Appendix~\ref{subsec: Training acceleration details}.

\section{Experiments}
\subsection{Experimental setup}\label{ExpSetup}

\textbf{Models and Data.} We employ open-source LLMs available on the HuggingFace Model Hub~\cite{wolf2019huggingface}, specifically Llama-3, Mistral-7B-v0.1, and Llama-2 model families. In all experiments, we train models on the open-source dataset RedPajama~\cite{weber2024redpajama}, as in prior VQ works. 

\textbf{Evaluation.} We adopt a standard benchmarking framework to compare our models with existing approaches. We report perplexity on the \mbox{WikiText-2~\cite{merity2016pointer}} test set and the validation set of the C4 dataset~\cite{raffel2020texttotext}. For perplexity, we use models’ native sequence length, in particular 4096 tokens for Llama2 and 8192 tokens for Llama3 and Mistral-7B-v0.1. We also provide zero-shot accuracy on WinoGrande~\cite{sakaguchi2021winogrande}, PiQA~\cite{bisk2020piqa}, HellaSwag~\cite{zellers2019hellaswag}, ARC-easy and ARC-challenge Question Answering (QA) benchmarks~\cite{clark2018arc}, using the Language Model Evaluation Harness library~\cite{eval-harness} for evaluation.

\textbf{Training parameters.}
During both Stage-1 and Stage-2 models are trained using the standard Adam~\cite{DBLP:journals/corr/KingmaB14} optimizer, for Stage-1 we also use Cosine Annealing scheduler~\cite{loshchilov2017sgdr}. To support reproducibility, we provide all relevant hyperparameters -- such as calibration data size, the number of epochs, learning rates, and index update steps -- for both stages in Appendix~\ref{subsec: Experimental Setup details}. For Stage-2, we use the same hyperparameters as in prior methods.

\subsection{Experimental results}
\begin{table}[]

\centering
\setlength\tabcolsep{2pt}

\caption{Evaluation of quantized Llama2-7B/70B, Llama3-8B/70B, and Mistral-7B-v0.1 models. The table reports perplexity measured on the WikiText-2 test set (W2) and C4 validation set, and accuracy for 0-shot question answering benchmarks: WinoGrande (WG), PiQA (PQ), HellaSwag (HS), ARC-easy (AE), ARC-challenge (AC) and QA Average with and without HellaSwag benchmark. As in prior works, we use a context length of 4096 for the Llama2 models and a context length of 8192 for both Mistral-7B-v0.1 and Llama3 models.}
\label{table:main-all-models}

\begin{tabular}{p{7pt}r|c|c|c|c|c|c|c|c|c}
    \hline
    &
    \bf{Method} &
    \bf{Bits} &
    \bf{W2$\downarrow$} &
    \bf{C4$\downarrow$} &
    \bf{WG$\uparrow$} &
    \bf{PQ$\uparrow$} &
    \bf{HS $\uparrow$} &
    \bf{AE$\uparrow$} &
    \bf{AC$\uparrow$}   &
    {\bf{QA Avg.$\uparrow$}} \\
    \hline
\multirow{8}{*}{\begin{turn}{90}\bf{Llama3-8B}\end{turn}} &
    FP16 &
    16 &
    5.54 &
    7.10 &
    73.1 &
    79.6 &
    60.2 &
    80.1 &
    50.3 &
    68.7\,/\,70.8 \\

    &
    QuIP\# &
    2.00 &
    7.84 &
    9.06 &
    68.2 &
    75.6 &
    - &
    72.9 &
    39.2 &
    -\,/\,64.0 \\

    &
    VPTQ &
    2.08  &
    - &
    - &
    65.9 &
    75.1 &
    52.2 &
    71.0 &
    36.9 &
    60.2\,/\,62.2 \\

    &
    QTIP &
    2.00 &
    7.33 &
    8.62 &
    \underline{70.7} &
    \underline{77.6} &
    - &
    75.2 &
    \underline{44.2} &
    -\,/\,\underline{66.9} \\

    &
    PV-Tuning &
    2.01 &
    \textbf{6.99} &
    \textbf{8.29} &
    69.9 &
    \textbf{77.8} &
    \underline{55.5} &
    \underline{75.8} &
    42.8 &
    \underline{64.4}\,/\,66.6 \\
    & \cellcolor{gray!15}TVQ+BasicVQ &
    \cellcolor{gray!15}2.02 &
    \cellcolor{gray!15}7.59 &
    \cellcolor{gray!15}8.86 &
    \cellcolor{gray!15}\textbf{71.5} &
    \cellcolor{gray!15}76.6 &
    \cellcolor{gray!15}54.3 &
    \cellcolor{gray!15}73.5 &
    \cellcolor{gray!15}41.3 &
    \cellcolor{gray!15}63.4\,/\,65.7 \\

    &\cellcolor{gray!15}TVQ+QuIP\# &
    \cellcolor{gray!15}2.00 &
    \cellcolor{gray!15}7.56 &
    \cellcolor{gray!15}8.82 &
    \cellcolor{gray!15}69.1 &
    \cellcolor{gray!15}76.9 &
    \cellcolor{gray!15}54.3 &
    \cellcolor{gray!15}74.4 &
    \cellcolor{gray!15}41.8 &
    \cellcolor{gray!15}63.3\,/\,65.6 \\

    &
    \cellcolor{gray!15}TVQ+QTIP &
    \cellcolor{gray!15}2.00 &
    \cellcolor{gray!15}\underline{7.13} &
    \cellcolor{gray!15}\underline{8.43} &
    \cellcolor{gray!15}70.2 &
    \cellcolor{gray!15}77.5 &
    \cellcolor{gray!15}\textbf{55.8} &
    \cellcolor{gray!15}\textbf{76.3} &
    \cellcolor{gray!15}\textbf{45.3} &
    \cellcolor{gray!15}\textbf{65.0}\,/\,\textbf{67.3} \\

\hline

\multirow{8}{*}{\begin{turn}{90}\bf{Llama3-70B}\end{turn}} &
    FP16 &
    16 &
    2.59 &
    5.78 &
    80.9 &
    82.5 &
    66.3 &
    86.7 &
    60.4 &
    75.4\,/\,77.6 \\

    &
    QuIP\# &
    2.00 &
    5.77 &
    7.46 &
    68.9 &
    54.7 &
    - &
    32.2 &
    18.3 &
    -\,/\,43.5 \\

    &
    VPTQ &
     2.08 &
    - &
    - &
    77.9 &
    80.4 &
    61.7 &
    81.8 &
    52.5 &
    70.9\,/\,73.2 \\

    &
    QTIP &
    2.00 &
    4.97 &
    6.80 &
    72.6 &
    57.1 &
    - &
    35.2 &
    28.0 &
    -\,/\,48.2 \\

    &
    PV-Tuning &
    2.07 &
    \underline{4.57} &
    \textbf{6.56} &
    \underline{78.1} &
    79.2 &
    \underline{63.9} &
    80.2 &
    50.8 &
    70.4\,/\,72.1 \\

    &
    \cellcolor{gray!15}TVQ+BasicVQ &
    \cellcolor{gray!15}2.07 &
    \cellcolor{gray!15}4.74 &
    \cellcolor{gray!15}6.77 &
    \cellcolor{gray!15}76.3 &
    \cellcolor{gray!15}78.2 &
    \cellcolor{gray!15}63.2 &
    \cellcolor{gray!15}79.1 &
    \cellcolor{gray!15}48.0 &
    \cellcolor{gray!15}69.0\,/\,70.4 \\

    &
    \cellcolor{gray!15}TVQ+QuIP\# &
    \cellcolor{gray!15}2.00 &
    \cellcolor{gray!15}5.33 &
    \cellcolor{gray!15}7.02 &
    \cellcolor{gray!15}75.5 &
    \cellcolor{gray!15}\underline{81.1} &
    \cellcolor{gray!15}62.0 &
    \cellcolor{gray!15}\textbf{83.7} &
    \cellcolor{gray!15}\underline{54.6} &
    \cellcolor{gray!15}\underline{71.4}\,/\,\underline{73.7} \\

    &
    \cellcolor{gray!15}TVQ+QTIP &
    \cellcolor{gray!15}\cellcolor{gray!15}2.00  &
    \cellcolor{gray!15}\textbf{4.48} &
    \cellcolor{gray!15}\underline{6.59} &
    \cellcolor{gray!15}\textbf{78.7} &
    \cellcolor{gray!15}\textbf{82.1} &
    \cellcolor{gray!15}\textbf{64.3} &
    \cellcolor{gray!15}\underline{83.2} &
    \cellcolor{gray!15}\textbf{55.9} &
    \cellcolor{gray!15}\textbf{72.8}\,/\,\textbf{75.0} \\

\hline

\multirow{7}{*}{\begin{turn}{90}\bf{Mistral-7B-v0.1}\end{turn}} &
    FP16 &
    16 &
    4.77 &
    5.71 &
    73.6 &
    80.5 &
    61.2 &
    78.9 &
    49.2  &
    68.7\,/\,70.6 \\

    &
    AQLM &
    2.01 &
    5.76 &
    6.60 &
    68.7 &
    77.6 &
    56.4 &
    73.3 &
    42.7 &
    63.7\,/\,65.6 \\

    &
    VPTQ &
    2.04 &
    5.64 &
    6.43 &
    68.7 &
    77.9 &
    56.1 &
    72.2 &
    41.1 &
    63.2\,/\,65.0 \\

    &
    PV-Tuning &
    2.01 &
    \textbf{5.29} &
    \textbf{6.17} &
    \textbf{72.8} &
    \underline{79.1} &
    \underline{58.2} &
    77.4 &
    44.2 &
    \underline{66.3}\,/\,\underline{68.4} \\

    &
    \cellcolor{gray!15}TVQ+BasicVQ &
    \cellcolor{gray!15}2.01 &
    \cellcolor{gray!15}5.59 &
    \cellcolor{gray!15}6.60 &
    \cellcolor{gray!15}\underline{71.4} &
    \cellcolor{gray!15}78.5 &
    \cellcolor{gray!15}56.9 &
    \cellcolor{gray!15}77.3 &
    \cellcolor{gray!15}43.2 &
    \cellcolor{gray!15}65.5\,/\,67.6 \\

    &
    \cellcolor{gray!15}TVQ+QuIP\# &
    \cellcolor{gray!15}2.00 &
    \cellcolor{gray!15}5.52 &
    \cellcolor{gray!15}6.53 &
    \cellcolor{gray!15}71.2 &
    \cellcolor{gray!15}78.4 &
    \cellcolor{gray!15}57.9 &
    \cellcolor{gray!15}\underline{77.8} &
    \cellcolor{gray!15}\underline{46.3} &
    \cellcolor{gray!15}\underline{66.3}\,/\,\underline{68.4} \\

    &
    \cellcolor{gray!15}TVQ+QTIP &
    \cellcolor{gray!15}2.00 &
    \cellcolor{gray!15}\underline{5.38} &
    \cellcolor{gray!15}\underline{6.39} &
    \cellcolor{gray!15}\underline{71.4} &
    \cellcolor{gray!15}\textbf{79.2} &
    \cellcolor{gray!15}\textbf{58.4} &
    \cellcolor{gray!15}\textbf{78.1} &
    \cellcolor{gray!15}\textbf{47.0} &
    \cellcolor{gray!15}\textbf{66.8}\,/\,\textbf{68.9} \\

\hline

\multirow{8}{*}{\begin{turn}{90}\bf{Llama2-7B}\end{turn}} &
    FP16 &
    16 &
    5.12 &
    6.63 &
    67.3 &
    78.5 &
    56.7 &
    69.3 &
    40.0 &
    62.4\,/\,63.8 \\

    &
    AQLM &
    2.02 &
    6.14 &
    8.09 &
    65.7 &
    76.0 &
    51.8 &
    63.4 &
    34.4 &
    58.3\,/\,59.9 \\

    &
    QuIP\# &
    2.00 &
    6.19 &
    8.16 &
    64.9 &
    75.1 &
    -  &
    64.6 &
    34.6 &
    -\,/\,59.8 \\

    &
    VPTQ &
    2.02 &
    6.13 &
    8.07 &
    64.3 &
    75.2 &
    52.1 &
    63.8 &
    35.2 &
    58.1\,/\,59.6 \\

    &
    QTIP &
    2.00 &
    5.86 &
    7.73 &
    64.7 &
    75.9 &
    - & 
    65.6 &
    35.7 &
    -\,/\,60.5\\

    &
    \cellcolor{gray!15}TVQ+BasicVQ &
    \cellcolor{gray!15}2.02 &
    \cellcolor{gray!15}6.11 &
    \cellcolor{gray!15}7.99 &
    \cellcolor{gray!15}65.4 &
    \cellcolor{gray!15}76.3 &
    \cellcolor{gray!15}\underline{52.8} &
    \cellcolor{gray!15}\textbf{67.7} &
    \cellcolor{gray!15}\underline{37.5} &
    \cellcolor{gray!15}\underline{59.9}\,/\,\textbf{61.7} \\

    &
    \cellcolor{gray!15}TVQ+QuIP\# &
    \cellcolor{gray!15}2.00 &
    \cellcolor{gray!15}6.01 &
    \cellcolor{gray!15}7.96 &
    \cellcolor{gray!15}\textbf{66.5} &
    \cellcolor{gray!15}\underline{76.9} &
    \cellcolor{gray!15}52.6 &
    \cellcolor{gray!15}65.1 &
    \cellcolor{gray!15}\textbf{37.6} &
    \cellcolor{gray!15}59.7\,/\,\underline{61.5} \\

    &
    \cellcolor{gray!15}TVQ+QTIP &
    \cellcolor{gray!15}2.00 &
    \cellcolor{gray!15}\underline{5.85} &
    \cellcolor{gray!15}\underline{7.72} &
    \cellcolor{gray!15}\underline{65.8} &
    \cellcolor{gray!15}\textbf{77.2} &
    \cellcolor{gray!15}\textbf{53.5} &
    \cellcolor{gray!15}\underline{66.9} &
    \cellcolor{gray!15}36.9 &
    \cellcolor{gray!15}\textbf{60.1}\,/\,\textbf{61.7} \\

\hline

\multirow{8}{*}{\begin{turn}{90}\bf{Llama2-70B}\end{turn}} &
    FP16 &
    16 & 
    3.12 &
    4.97 &
    77.0 &
    81.1 &
    64.0 &
    77.7 &
    51.1 &
    70.2\,/\,71.7 \\

    &
    AQLM &
    2.07 &
    3.83 &
    5.62 &
    74.4 &
    \underline{80.9} &
    62.2 &
    74.6 &
    \underline{49.0} &
    68.2\,/\,69.7 \\

    &
    QuIP\# &
    2.00 & 
    3.91 &
    5.71 &
    \underline{75.9} &
    80.3 &
    62.5 &
    \underline{77.3} &
    48.7 &
    \underline{68.9}\,/\,\underline{70.6} \\

    &
    VPTQ &
    2.07 &
    3.93 &
    5.72 &
    75.0 &
    80.3 &
    \textbf{63.0} &
    77.1 &
    47.7 &
    68.6\,/\,70.0 \\

    &
    QTIP &
    2.00 &
    \textbf{3.70} &
    \textbf{5.48} &
    75.1 &
    80.2 &
    - &
    76.3 &
    48.0 &
    -\,/\,69.9 \\

    &
    \cellcolor{gray!15}TVQ+BasicVQ &
    \cellcolor{gray!15}2.07 &  
    \cellcolor{gray!15}3.88 & 5.77 &
    \cellcolor{gray!15}\textbf{77.2}  &
    \cellcolor{gray!15}80.8 &
    \cellcolor{gray!15}62.5  &
    \cellcolor{gray!15}\textbf{77.6}   &
    \cellcolor{gray!15}\textbf{49.2}    &
    \cellcolor{gray!15}\textbf{69.5}\,/\,\textbf{71.2} \\

    & \cellcolor{gray!15}TVQ+QuIP\# & 
    % TVQ+QuIP\# & 
    \cellcolor{gray!15}2.00 &  
    \cellcolor{gray!15}3.96 & 5.81 &
    \cellcolor{gray!15}75.5  &
    \cellcolor{gray!15}80.5  &
    \cellcolor{gray!15}62.1   &
    \cellcolor{gray!15}76.4   &
    \cellcolor{gray!15}48.4    &
    \cellcolor{gray!15}68.6\,/\,70.2 \\

    & \cellcolor{gray!15}TVQ+QTIP &
    % TVQ+QTIP & 
    \cellcolor{gray!15}2.00 &  
    \cellcolor{gray!15}\underline{3.72}  & \underline{5.50} &
    \cellcolor{gray!15}74.6   &
    \cellcolor{gray!15}\textbf{81.3}  &
    \cellcolor{gray!15}\underline{62.9}   &
    \cellcolor{gray!15}77.0   &
    \cellcolor{gray!15}48.2  &
    \cellcolor{gray!15}68.8\,/\,70.3 \\
\hline

\end{tabular}

\end{table}

We present our experimental results in Table~\ref{table:main-all-models}. Our analysis focuses on the QA accuracy averaged across multiple benchmarks: WinoGrande, PiQA, HellaSwag, Arc Easy, and Arc Challenge. When comparing TVQ with QuIP\# and QTIP, we exclude HellaSwag from the averaging calculation, as these methods do not provide metrics for this benchmark.

\textbf{Comparision of optimization approaches.}
To fairly compare TVQ with previous optimization approaches, we consider that both the VQ scheme and the optimization method contribute to the final quality. Therefore, for several previous works, we replaced their optimization method with TVQ, maintaining their original VQ scheme. In this way, we isolated performance improvements attributable to our TVQ. In particular, we applied TVQ with the BasicVQ scheme defined in Section~\ref{subsec: Vector Quantization Background}. These models, which we refer to as TVQ+BasicVQ, are architecturally identical to those used in AQLM. We also performed experiments with the quantization schemes proposed in QuIP\# and QTIP. The resulting experimental configurations are referred to as TVQ+QuIP\# and TVQ+QTIP, respectively.

By comparing results in pairs of VQ models that differ only in optimization method (TVQ+BasicVQ vs AQLM, TVQ+QuIP\# vs QuIP\#, TVQ+QTIP vs QTIP), we observe that TVQ consistently outperforms the layer-wise optimization strategies initially employed in AQLM, QuIP\#, and QTIP methods. 
For instance, TVQ outperforms the beam search used in AQLM across all models by 1.5\% on average, reducing the average compression-related accuracy drop from 3.7\% to 2.2\%. 

TVQ also improves results for the QuIP\# VQ scheme by 0.9\% on average across all models except for LLama3-70B. For LLama3-70B, the improvement is particularly substantial, where the original QuIP\# algorithm achieves only 43.5\% accuracy, while TVQ+QuIP\# reaches 73.7\%. This considerable gap, undetected by perplexity metrics, also reveals the limited predictive value of perplexity for assessing quantization impact on downstream task performance in practical applications.

All TVQ+QTIP models achieve higher QA accuracy than their corresponding baseline QTIP models across all tested models. The combination of QTIP's advanced quantization scheme with TVQ's optimization approach yields the highest scores across multiple model architectures, establishing TVQ+QTIP as the leading configuration in our evaluation.

\textbf{State-of-the-art performance.} Models optimized with TVQ achieve state-of-the-art performance for all evaluated architectures. On Llama2-70B, TVQ achieves near-lossless results with only 0.7\% performance degradation relative to the original model, improving upon the previous best result from QuIP\# (1.3\% accuracy degradation). For Llama2-7B, TVQ outperforms all other methods by 1-2\%. On Llama3-70B, TVQ shows a 1.9\% improvement, reducing the performance gap to the full-precision baseline from 4.5\% to 2.6\%. Unlike existing approaches, where no single method consistently leads (e.g., VPTQ performs best on Llama3-70B, PV-tuning on Mistral-7B-v0.1, and QTIP on Llama3-8B), TVQ achieves a new state-of-the-art quality in low-bit PTQ for all considered architectures.

\subsection{Training efficiency}\label{subsec: Training efficiency}
While TVQ is primarily focused on the quality and simplicity of implementation, it also operates within a moderate computational budget, as shown in Table \ref{table: Training Efficiency}. TVQ+QuIP\# and TVQ+BasicVQ are equally compute-consuming as QuIP\# and AQLM correspondingly, while achieving better quality.

\begin{table}[H]

\centering

\setlength\tabcolsep{3pt}

\caption{Quantization computational costs of 2-bit Llama2-70B. The table reports the number of GPUs hours measured using a single NVIDIA 80GB Ampere100 GPU. 
As PV-tunning is applied on top of Stage-1 methods, we refer to its Stage-1 GPU hours as X. Note that the training time is significantly shorter since most methods, including TVQ, can be parallelized across multiple GPUs.}
\label{table: Training Efficiency}

\begin{tabular}{l r c l}

\hline
    \textbf{Method} & \textbf{Stage-1} & \textbf{Stage-2} & \textbf{Total} \\
\hline
   AQLM & 250 & 50 & 300  \\
   VPTQ & 74 & 50 & 124  \\
   Quip\# & 40 & 50 & 90  \\
   QTIP & 50 & 50 & 100  \\
   PV-tuning & X & 1500 & >1500  \\
   \rowcolor{gray!15}
   TVQ+Quip\# & 54 & 50 & 104  \\
   \rowcolor{gray!15}
   TVQ+BasicVQ & 212 & 50 & 262  \\
   \rowcolor{gray!15}
   TVQ+QTIP & 390 & 50 & 440  \\
\hline

\end{tabular}
\end{table}

TVQ+QTIP reaches better quality than PV-tuning, which demands a huge computational budget ($\approx$1,500 GPU-hours for Stage-2 alone) and requires an 8 Ampere100 GPU node. In contrast, our pipeline completes within 100-400 total GPU-hours using only 2–4 Ampere100 GPUs — the same scale as other efficient methods — while achieving better model quality.
This demonstrates that our method attains superior quantization performance with reasonable computational cost, making it practical and applicable in real-world settings.

\begin{table}[H]

\centering
\setlength\tabcolsep{3pt}
\caption{Ablation study of the scheduled index matrix updates for TVQ+QTIP. The table reports results after Stage-1 optimization only.
The training time is measured using 8 NVIDIA 32GB Volta100 GPU. Note that training time is significantly reduced when training on modern GPU architectures, such as Ampere or Hopper.
}
\label{table: ablation TVQ QTIP different upd_step}

\begin{tabular}{cc cc cc cc}
\hline
\multirow{2}{*}{\textbf{\textbf{$\mathbf{S}_{\text{upd}}$}}}
  & \multirow{2}{*}{\makecell{\textbf{Stage-1} \\ \textbf{time}}}
  & \multicolumn{2}{c}{\textbf{Llama2-7B}} 
  & \multicolumn{2}{c}{\textbf{Llama3-8B}} 
  & \multicolumn{2}{c}{\textbf{Mistral-7B}} \\
  
 &
 & W2$\downarrow$ & \hspace*{2pt} QA$\uparrow$ 
 & W2$\downarrow$ & \hspace*{2pt} QA$\uparrow$ 
 & W2$\downarrow$ & QA$\uparrow$ \\
\hline
8  & $\approx$32h & \hspace*{2pt} 6.15 & \hspace*{2pt} 59.0 & \hspace*{2pt} 8.45 & \hspace*{2pt} 63.3 &  \hspace*{2pt} 5.72 & 65.8 \\
\rowcolor{gray!15}
16 & $\approx$21h & \hspace*{2pt} 6.17 & \hspace*{2pt} 58.9 & \hspace*{2pt} 8.60 & \hspace*{2pt} 62.2 & \hspace*{2pt} 5.71 & 66.0 \\
32 & $\approx$15h & \hspace*{2pt} 6.22 & \hspace*{2pt} 58.8 & \hspace*{2pt} 8.68 & \hspace*{2pt} 62.0 & \hspace*{2pt} 5.74 & 65.7 \\
64 & $\approx$14h & \hspace*{2pt} \textgreater{}100 & \hspace*{2pt} \textless{}10 & \hspace*{2pt} \textgreater{}100 & \hspace*{2pt} \textless{}10 &\hspace*{2pt}  \textgreater{}100 & \textless{}10 \\
\hline
\end{tabular}

\end{table}

\subsection{Ablation study of Scheduled Index Updates}\label{subsec: ablation scheduled updates}

For both TVQ+BasicVQ and TVQ+QTIP experiments, the indices function $\displaystyle{I(W, C)}$ computation constitutes a significant part of the overall computational cost. To address this bottleneck, we implement a strategy that performs index matrices updates every $\displaystyle{S_{\text{upd}}}$ optimization steps.

\begin{table}[H]

\centering
\setlength\tabcolsep{3pt}

\caption{Ablation study of the scheduled index matrix updates for TVQ+BasicVQ experiments. The table reports results after Stage-1 optimization only.
% perplexity measured on the WikiText-2 test set (W2) and average accuracy for zero-shot question answering benchmarks: WinoGrande, PiQA, HellaSwag, ARC-easy, and ARC-challenges.
The training time is measured using 2 NVIDIA 80GB Ampere100 GPUs.}
\label{table: updateFreq_quality_trainingTime}

\begin{tabular}{c c c c c}
\hline 
\multirow{2}{*}{\textbf{$\mathbf{S}_{\text{upd}}$}} & \textbf{Stage-1} & \multicolumn{2}{c}{\makecell{\textbf{Llama2-70B}}}  \\
% \cline{2-3}
 & \textbf{time} & W2$\downarrow$ & QA$\uparrow$ \\
\hline
\rowcolor{gray!15}
1 & 51h & 4.11 & 68.2  \\
2 & 33h & 4.10 & 68.1  \\
3 & 24h & 4.11 & 68.1 \\
4 & 19h & 4.15 & 67.8  \\
\hline

\end{tabular}

\end{table}

For TVQ+QTIP, we evaluate how model quality depends on $\displaystyle{S_{\text{upd}}}$ in Table~\ref{table: ablation TVQ QTIP different upd_step}. Experimental results demonstrate that setting $\displaystyle{S_{\text{upd}}}$ to $8$, $16$, and $32$, yields comparable quality metrics, while increasing $\displaystyle{S_{\text{upd}}}$ to $64$ leads to a significant quality degradation. Interestingly, experiments with $S_{\text{upd}}$ equal to $32$ and $64$ exhibit nearly identical training times, as the index recomputation overhead becomes negligible relative to the total training cost. Based on these observations, we conclude that setting $\displaystyle{S_{\text{upd}}}$ to $16$ provides an optimal trade-off between quality and computational efficiency for the majority of TVQ+QTIP experiments.

For TVQ+BasicVQ experiments, we set $\displaystyle{S_{\text{upd}}}$ to 1 as the default, meaning that indices are recomputed at each optimization step. This configuration prioritizes model quality over computational efficiency. We analyze the impact of $\displaystyle{S_{\text{upd}}}$ on both quality metrics and training time in Table~\ref{table: updateFreq_quality_trainingTime}. This table demonstrates that increasing $\displaystyle{S_{\text{upd}}}$ to 2 or 4 can accelerate experiments in practical scenarios, while incurring only a minor drop in quality.

\section{Conclusions}
In this work, we tackled the problem of vector assignment optimization in VQ models, introducing TVQ – an algorithm that enables gradient-based index optimization with arbitrary loss functions. Our experimental results confirm that TVQ consistently outperforms previous VQ optimization methods 
on popular LLM evaluation benchmarks. Moreover, TVQ is applicable to all existing vector quantization schemes and is straightforward to implement using popular deep learning frameworks. We believe that future research can leverage TVQ for index assignment optimization while advancing VQ architectures, further improving LLM efficiency.

\section{Limitations}
Several works, such as QTIP and AQLM, provide software tools for faster execution on modern hardware, such as CUDA kernels for NVIDIA GPUs. These works have shown that vector quantization not only reduces memory consumption but also significantly benefits throughput and latency. One limitation of our work is that we do not contribute to the further development of hardware-specific utilities for vector-quantized models. However, since TVQ preserves the structure of optimized models, they remain fully compatible with existing software tools and inherit the same efficiency and speedups as prior checkpoints.

\bibliographystyle{splncs04}
\bibliography{custom}

\appendix
\newpage
\section{Existing VQ schemes}\label{subsec: appendix vq schemes}
In this section, we summarize quantization schemes employed in related works. To introduce our framework, we formally define a VQ scheme in Section~\ref{subsec: vq schemes} as a pair  $\displaystyle{\left(\mathcal{C},\mathcal{I}\right)}$, where $\mathcal{C}$ denotes the codebook matrix search space and $\mathcal{I}$ denotes the index matrix search space. In Table~\ref{table:VQ_schemes_summary}, we specify such pairs that characterize the quantization schemes adopted in prior works.

These schemes are used in our experiments with TVQ for index optimization. In TVQ+BasicVQ, we use the same scheme as in AQLM and VPTQ, where codebook and index matrices are unconstrained. In TVQ+Quip\#, we use the scheme from QuIP\#, where the codebook matrix is restricted to a single E8P codebook, which remains fixed throughout the entire optimization process, while the index matrices are unconstrained. For details on the construction and properties of the E8P codebook, we refer the reader to the QuIP\# paper. In TVQ+QTIP, we employ the same scheme as QTIP, where hybrid (HYB) codebooks are constrained to those that can be constructed via bitwise operations on vectors drawn from a smaller, unconstrained codebook $C_s$. The index matrices are restricted to those that can be compressed using the bitshift trellis bit-packing algorithm. For a detailed explanation of the bitshift trellis algorithm and the HYB codebooks, we refer to the QTIP paper.

\section{TVQ implementation}\label{subsec: TVQ implementation}
In this section, we detail the TVQ implementation. Algorithm~\ref{alg:highlvel} presents the pseudocode for the TVQ Training Pipeline described in Section~\ref{subsec: training pipeline}. Listing~\ref{listing_classes} details our PyTorch implementation of the quantized weight classes used during Stage-1. In a pretrained model, the weights in the linear layers of Transformer blocks are replaced with one of the classes from Listing~\ref{listing_classes}. Class \lstinline|WeightTVQ| implements the core logic of the proposed TVQ method. Classes \lstinline!WeightTVQ_BasicVQ,! \lstinline|WeightTVQ_QTIP,| and \lstinline|WeightTVQ_QUIPsharp| implement corresponding quantization schemes.
\lstinline|IndicesUpdatesScheduler| is used for scheduled indices updates (see Section~\ref{subsec: implementation details} and Section~\ref{subsec: ablation scheduled updates}).

\section{Experimental Setup details}\label{subsec: Experimental Setup details}
In this section, we report the full set of hyperparameters employed at each stage of our pipeline. All experiments were conducted using the PyTorch library.

Table~\ref{table: stage 1 experiments parameters} provides the hyperparameters used for Stage-1. We used the Adam optimizer with default weight decay and beta values, along with the Cosine Annealing scheduler. All models were optimized using the reconstruction $L_1$ loss. The remaining hyperparameters varied depending on the specific configuration and are reported in the table. Table~\ref{table: stage 2 experiments parameters} provides the hyperparameters for Stage-2. During Stage-2, we optimize the KL divergence between the outputs of the quantized model and the original full-precision model. The parameters listed include learning rates, batch sizes, number of optimization steps, and other settings relevant to the fine-tuning process.

\begin{table*}[th!]
\begin{center}

\setlength\tabcolsep{3pt}

\caption{Summary of quantization schemes used in prior works. From the optimization perspective, each vector quantization scheme is formally defined by the set of permissible codebooks $\displaystyle{\mathcal{C}}$ and the set of permissible index matrices $\displaystyle{\mathcal{I}}$ (see Section~\ref{subsec: vq schemes}).}
\label{table:VQ_schemes_summary}

\begin{tabular}{c c c}
    \hline
    \multirow[c]{2}{*}{\bf{Method}} & \multicolumn{2}{l}{\Gape[1mm]{\hspace{35mm} \bf{VQ scheme}}} \\
    \cline{2-3}

    & \Gape[1mm]{\makecell{\bf{Codebook matrix}\\ \bf{constraints}}} & \makecell{\bf{Index matrix}\\ \bf{constraints}} \\
    \hline
    \Gape[1mm]{\makecell{AQLM,\\ VPTQ}}  
    & \makecell{$\mathcal{C} = \mathbb{R}^{n_c \times d} $ \\(unconstrained)} 
    & \makecell{$\mathcal{I} = \big\{ 0 \dots n_C - 1 \big\}^{d_{\text{out}}, \frac{d_{\text{in}}}{d}}$  \\ (unconstrained)} \\
    \hline
    \Gape[1mm]{QuIP\#} 
    & \makecell{$\mathcal{C} = \big\{ \text{E8P codebook} \big\}$} & \makecell{$\mathcal{I} = \big\{ 0 \dots n_C - 1 \big\}^{d_{\text{out}}, \frac{d_{\text{in}}}{d}}$  \\ (unconstrained)} \\
    \hline
     \Gape[9mm]{QTIP} 
    & \makecell{$\mathcal{C} = \big\{ C \mid C = \text{HYB}(C_s) \big\}$, \\ where $C_s \in \mathbb{R}^{N_s \times d}$ is \\ an unconstrained \\ smaller codebook \\ that fits into L1 GPU cache } 
    & \makecell{$\displaystyle{\mathcal{I} = \left\{I \big| i_k \equiv \frac{i_{k+1}}{{2^{(\lceil \log_2 (n_C) \rceil  - d \cdot K)}}} \mod 2^{d \cdot K}\right\}}$, \\ for all $\displaystyle{i_k \in i = vec(I)}$, \\ where $K$ is the average bitwidth \\ of the packed index matrix} \\
    \hline    

\end{tabular}

\end{center}
\end{table*}

\begin{table}[H]
\centering

\caption{Ablation study of the Approximate Nearest Neighbor search for the 2-bit Llama2-7B. $\displaystyle{N}$ denotes the number of Voronoi cells, $\displaystyle{K}$~--- the number of visited cells. The grey line highlights the selected setting. The table shows evaluation metrics for the 2-bit Llama2-7B model after Stage-1, without the fine-tuning Stage-2.}
\label{table: ablation ANN search}

\begin{tabular}{c c c c c c}

\hline
  \makecell{\bf{Search} \\ \bf{Accuracy}} & \bf{N} & \bf{K} & \makecell{\bf{Speedup}} & \bf{W2$\downarrow$} & \bf{C4$\downarrow$}\\
\hline
   99.9\,\%  &  256  &  16  &  $\times$5.0  &  6.70  &  8.70 \\
   99\,\%  &  1024  &  16  &  $\times$5.5  &  6.64  &  8.68  \\
  \rowcolor{gray!15}
   98\,\%  &  256  &  8  &  $\times$6.4  & 6.63  &  8.69  \\  
   95\,\%  &  512  &  8  &  $\times$6.5  & 6.68  &  8.73  \\  
   80\,\%  &  1024  &  4  &  $\times$6.7  &  6.84  &  8.92  \\ 
   60\,\%  &  1024  &  2  &  $\times$6.7  & 7.13  &  9.15  \\ 
\hline

\end{tabular}

\end{table}

\section{Ablation study of Approximate Nearest Neighbor Search}\label{sec: ANN ablation}

In TVQ+BasicVQ experiments, the index function $\displaystyle{I(W, C)}$, defined in Equation~\eqref{eq:I_from_W}, maps each vector of the trainable weight $\displaystyle{W}$ to the index of its Nearest-Neighbor in the codebook $\displaystyle{C}$. The exact Nearest-Neighbor search requires computing pairwise distances between all codebook vectors and trainable weight vectors, reaching trillions of distance calculations for modern LLMs. As a result, the exhaustive search presents a significant computational bottleneck.

To solve this problem, we adopt the Approximate Nearest-Neighbor (ANN) search via Voronoi cells, which is implemented in the open-source library FAISS. This approach groups codebook vectors into clusters named Voronoi cells. Then, for each vector of the trainable weight $\displaystyle{W}$, it finds several nearest cells and performs an exhaustive Nearest-Neighbor search only within these cells. Thus, a large number of distant vectors is excluded from the search space, and a naive exhaustive Nearest-Neighbor search is only performed with the remainder.

We analyze the impact of the Approximate Nearest-Neighbor search parameters on the final model quality. We conduct experiments, varying the parameters that affect the search quality, particularly the number of Voronoi cells $\displaystyle{N}$ and the number of cells for exhaustive search $\displaystyle{K}$. Search quality is evaluated using accuracy, which is defined as the percentage of queries where the approximate search matches the exact search. Additionally, we measure the speedup by comparing the approximate search against the exact search implemented on a single NVIDIA 32GB Volta100 GPU. As shown in Table~\ref{table: ablation ANN search}, the approximate search that reaches 99.99\% of accuracy is five times faster than the exact search. The model quality remains unaffected even when search accuracy drops to 95\%, and the decline is minimal even at 80\% accuracy. These findings suggest that opting for ANN search over exact search is unlikely to degrade the model quality.

\section{Distribution of the number of VQ parameters}\label{subsec: Bits/weight details}
Since different VQ schemes employ distinct configuration settings — such as codebook sizes, vector dimensionality, and storage strategies — the distribution of the number of VQ parameters and their storage requirements varies accordingly.

In the BasicVQ scheme, all weights in both index matrices and codebooks are subject to optimization and are retained for inference. For example, in 2-bit-per-weight quantization of the 7–8B models, $\approx$1.86 bits per weight contain information about the index matrices, while only $\approx$0.16 bits per weight correspond to the codebooks. For larger 70B models, index matrices constitute an even larger fraction of the total weights ($\approx$2 bits per weight for index matrices vs. $\approx$0.07 bits per weight for codebooks). Detailed estimations of the Basic-VQ model size are provided in the AQLM paper, which we refer to. In the QuIP\# and QTIP schemes, codebooks contain fewer than 0.01 bits per weight of adjustable parameters, while the index matrices contain approximately 2 adjustable bits per weight.

These observations conclude that the largest part of the parameters resides in index matrices, underscoring the importance of their optimization.

\begin{table}[H]

\centering

\setlength\tabcolsep{4pt}

\caption{Index matrix update time in seconds, measured on a single NVIDIA 80GB Ampere100 GPU ($\approx$2 bits VQ, $\displaystyle{W \in \mathbb{R}^{8192\times 8192}}$; $C \in \mathbb{R}^{2^{16}\times8}$ for Quip\# and BasicVQ, $C \in \mathbb{R}^{2^{16}\times2}$ for QTIP).}
\label{table: Training Efficiency 2}

\begin{tabular}{lcc}

\hline

    \textbf{Method} & \makecell{\textbf{Index update} \\ \textbf{algorithm}} & \makecell{\textbf{Update time} \\ \textbf{(sec)}} \\
\hline
   TVQ+Quip\# & E8-lattice NN search & 0.236 ± 0.001  \\
   TVQ+BasicVQ & ANN search & 1.31 ± 0.02  \\
   TVQ+QTIP & Viterbi search & 41.8 ± 0.2  \\
\hline

\end{tabular}

\end{table}

\begin{algorithm}[H]
\caption{Training Pipeline}
\label{alg:highlvel}

\textbf{Input}: pretrained model \texttt{fp\_model}, input data $\mathbf{D}$, weight class \texttt{WeightTVQ\_Class},  indices recomputation step \texttt{upd\_step}, number of optimization steps in Stage-1 $T_{1}$, Stage-2 $T_{2}$ \\
\textbf{Output}: vector-quantized model \texttt{vq\_model}

\begin{algorithmic}[1]

\STATE \textit{\#  Create vector quantized model}
\STATE \texttt{vq\_model} = \texttt{copy}(\texttt{fp\_model})
\FOR{$\texttt{layer} \in \texttt{linear\_layers}(\texttt{vq\_model})$}
    \STATE $\texttt{layer.weight} = \texttt{WeightTVQ\_Class}($ \\
    $\texttt{  layer.weight}, \texttt{upd\_step}$ \\
    $)$
\ENDFOR

\STATE \textit{\#  Block-Wise PTQ (Stage-1)}
\STATE $\mathbf{X} = \texttt{fp\_model.input\_embeddings}(\mathbf{D})$
\STATE $\mathbf{{X}^q} = \texttt{vq\_model.input\_embeddings}(\mathbf{D})$ \text

\FOR{$i = 1$ \TO \texttt{model.num\_blocks}} 
    \STATE $\texttt{fp\_block}$ = $\texttt{fp\_model.get\_block}(i)$
    \STATE $\texttt{vq\_block}$ = $\texttt{vq\_model.get\_block}(i)$
    
    \STATE \textit{\#  Optimize block}
    \FOR{$j = 1$ \TO $T_{1}$}
        \STATE $\mathcal{L}\!=\!\|\!\texttt{fp\_block}(\mathbf{X_{k}}\!)-\texttt{vq\_block}(\mathbf{{X}^q_{k}})\!\|_1$
        \STATE \textit{\footnotesize \# $k$ - batch index at iteration $j$}
        \STATE {Calculate gradients of $\mathcal{L}$}
        \STATE {Update W and C (optionally) with Adam}
    \ENDFOR

    \STATE \textit{\# Collect input data for the next block}
    \STATE $\mathbf{X} = \texttt{fp\_block}(\mathbf{X})$
    \STATE  $\mathbf{{X}^q} = \texttt{vq\_block}(\mathbf{{X}^q})$
\ENDFOR

\STATE \textit{\# End-to-End Fine-tuning (Stage-2)}

\FOR{$j = 1$ \TO $T_{2}$}
    \STATE $\mathcal{L} = \text{KL}(\texttt{vq\_model}(\mathbf{D_k}), \texttt{fp\_model}(\mathbf{D_k}))$
    \STATE \textit{\footnotesize \# $k$ - batch index at iteration $j$}
    \STATE {Calculate gradients of $\mathcal{L}$}
    \STATE {Update trainable parameters with Adam}
\ENDFOR

\end{algorithmic}
\end{algorithm}

\begin{table*}[tb!]
\centering

\caption{Training parameters used in Stage-1: input data sequence length $\left(\text{SL}\right)$, number of sequences $\displaystyle{\left(\text{SN}\right)}$, number of training epochs $\displaystyle{\left(\text{E}\right)}$, batch size $\displaystyle{\left(\text{BS}\right)}$, parameters of the Approximate Nearest-Neighbor search $\displaystyle{\left(\text{ANNs}, \text{the number of Voronoi cells} / \text{the number of visited cells}\right)}$, indices recomputation step  $\displaystyle{\left(\mathbf{S}_{\text{upd}}\right)}$ , learning rate for the codebooks $\displaystyle{\left(\text{LR}_C\right)}$, learning rate for the weights $\displaystyle{\left(\text{LR}_W\right)}$.}
\label{table: stage 1 experiments parameters}

\begin{tabular}{ll cccccccc}
    \hline
        \multirow{2}{*}{\bf{Model}} & 
        \multirow{2}{*}{\bf{Method}} & 
        \multirow{2}{*}{\bf{SL}} & 
        \multirow{2}{*}{\bf{SN}} & 
        \multirow{2}{*}{\bf{E}}& 
        \multirow{2}{*}{\bf{BS}} & 
        \multirow{2}{*}{\bf{ANNs}} & 
        \multirow{2}{*}{\bf{$\mathbf{S}_{\text{upd}}$}} & 
        \multirow{2}{*}{\bf{$\text{LR}_C$}} & 
        \multirow{2}{*}{\bf{$\text{LR}_W$}} \\
    & & & & & & & & & \\
    \hline
        ~ & TVQ+BasicVQ & 4096 & 2048 & 2 & 8 & 256/8 & 1 & 5e-5 & 5e-5    \\ 
        Llama2-7B & TVQ+QuIP\# & 4096 & 2048 & 2 & 8 & - & 1 & - & 1e-5    \\ 
        ~ & TVQ+QTIP & 4096 & 2048 & 2 & 8 & - & 16 & - & 1e-5    \\ \hline
        ~ & TVQ+BasicVQ & 4096 & 2048 & 1 & 8 & 128/8 & 1 & 5e-5 & 5e-5    \\ 
        Llama2-70B & TVQ+QuIP\# & 4096 & 2048 & 1 & 8 & - & 2 & - & 1e-5    \\ 
        ~ & TVQ+QTIP & 4096 & 2048 & 1 & 8 & - & 16 & - & 1e-5    \\ \hline
        ~ & TVQ+BasicVQ & 8192 & 1024 & 2 & 4 & 256/8 & 1 & 1e-4 & 2e-5    \\ 
        Mistral-7B-v0.1 & TVQ+QuIP\# & 8192 & 1024 & 2 & 4 & - & 1 & - & 6e-6    \\ 
        ~ & TVQ+QTIP & 8192 & 1024 & 2 & 4 & - & 16 & - & 2e-6    \\ \hline
        ~ & TVQ+BasicVQ & 4096 & 2048 & 4 & 8 & 256/8 & 1 & 3e-4 & 1e-4    \\ 
        Llama3-8B & TVQ+QuIP\# & 4096 & 2048 & 2 & 8 & - & 1 & - & 5e-6    \\ 
        ~ & TVQ+QTIP & 4096 & 2048 & 4 & 8 & - & 8 & - & 4e-6    \\ \hline
        ~ & TVQ+BasicVQ & 4096 & 2048 & 1 & 8 & 128/8 & 4 & 5e-5 & 5e-5    \\ 
        Llama3-70B & TVQ+QuIP\# & 4096 & 2048 & 1 & 8 & - & 2 & - & 5e-6    \\ 
        ~ & TVQ+QTIP & 4096 & 2048 & 1 & 8 & - & 16 & - & 5e-6    \\ \hline
    \end{tabular}

\end{table*}

\begin{table*}[tb!]
\centering
\caption{Training parameters used in the Stage-2 of optimization: input data sequence length $\displaystyle{\left(\text{SL}\right)}$, number of sequences $\displaystyle{\left(\text{SN}\right)}$, batch size $\displaystyle{\left(\text{BS}\right)}$, learning rate for the codebooks $\displaystyle{\left(\text{LR}_{C}\right)}$, learning rate for the incoherence processing parameters $\displaystyle{\left(\text{LR}_{IP}\right)}$, learning rate for the language modeling head $\displaystyle{\left(\text{LR}_{lm\_head}\right)}$. }
\label{table: stage 2 experiments parameters}

\begin{tabular}{l l c c c c c c c }
\hline
    \multirow{2}{*}{\bf{Model}} & 
    \multirow{2}{*}{\bf{Method}} & 
    \multirow{2}{*}{\bf{SL}} & 
    \multirow{2}{*}{\bf{SN}}& 
    \multirow{2}{*}{\bf{BS}} & 
    \multirow{2}{*}{\bf{$\text{LR}_C$}} & 
    \multirow{2}{*}{\bf{$\text{LR}_{IP}$}} & 
    \multirow{2}{*}{\bf{$\text{LR}_{lm\_head}$}} \\
& & & & & & & \\
\hline   
 ~ & TVQ+BasicVQ & 4096  & 4096 & 50 & 1e-5 & - & - \\ 
 Llama2-7B & TVQ+QuIP\# & 4096 & 4096 & 50 & - & 5e-4 & 5e-5   \\ 
 ~ & TVQ+QTIP & 4096  & 4096& 50 & 1e-3 & 5e-4 & 5e-5   \\ \hline
 ~ & TVQ+BasicVQ & 4096 & 4096& 50 & 1e-5 & - & - \\
 Llama2-70B & TVQ+QuIP\# & 4096  & 8192 & 50 & - & 5e-4 & 5e-5   \\ 
 ~ & TVQ+QTIP & 4096  & 8192& 50 & 1e-3 & 5e-4 & 5e-5   \\ \hline
 ~ & TVQ+BasicVQ & 8192  & 4096& 50 & 1e-5 & - & -   \\ 
 Mistral-7B-v0.1 & TVQ+QuIP\# & 8192  & 4096& 50 & - & 5e-4 & 5e-5   \\ 
 ~ & TVQ+QTIP & 8192  & 4096 & 50 & - & 5e-4 & 5e-5   \\ \hline
 ~ & TVQ+BasicVQ & 8192  & 4096 & 50 & 1e-5 & - & -   \\ 
 Llama3-8B & TVQ+QuIP\# & 8192  & 4096 & 50 & - & 5e-4 & 5e-5  \\ 
 ~ & TVQ+QTIP & 8192  & 8192&50 & 5e-3 & 5e-4 & 5e-5   \\ \hline
 ~ & TVQ+BasicVQ & 8192  & 1024& 50& 1e-5 & - & -   \\
 Llama3-70B & TVQ+QuIP\# & 8192  & 1024&50 & - & 5e-4 & 5e-5  \\ 
 ~ & TVQ+QTIP & 8192  & 1024& 50& 5e-3 & 5e-4 & 5e-5   \\ \hline
\end{tabular}

\end{table*}

\begin{listing}[H]%
\caption{PyTorch implementation of TVQ Weight Classes}%
\begin{lstlisting}[language=Python,
label={listing_classes}, basicstyle=\scriptsize
]
class IndicesUpdateScheduler:
  def __init__(self, upd_step=1):
    self.upd_step = upd_step
    self.step = 0

  def __next__(self):
    self.step += 1
    state = self.step % self.upd_step
    return state==0

class WeightTVQ(nn.Module):
  def __init__(self, W_fp, codebook_init, indices_fn, processing, upd_step):
    super().__init__()
    self.processing = processing
    self.W = processing.preprocess(W_fp)
    self.C = codebook_init
    self.indices_fn = indices_fn    
    self.I = self.indices_fn(self.W, self.C)
    self.idx_upd_scheduler = IndicesUpdateScheduler(upd_step)
    
  def forward(self):
    if next(self.idx_upd_scheduler):
      self.I = self.indices_fn(
        self.W, self.C
      )
    W_q = self.C[self.I]
    W_q = W_q + self.W - self.W.detach()
    return self.processing.postprocess(W_q)

  def backward(self, grad_output):
    # automatic by PyTorch

class WeightTVQ_BasicVQ(WeightTVQ):
  def __init__(self, W, upd_step):
    processing = per_channel_scaling
    codebook = kmeans(
        processing.preprocess(W)
    )
    indices_fn = approx_nn_search
    super().__init__(W, codebook, indices_fn, processing, upd_step)
    
class WeightTVQ_QTIP(WeightTVQ):
  def __init__(self, W, upd_step):
    processing = incoherence_processing
    codebook = qtip_hyb_codebook
    indices_fn = viterbi_search
    super().__init__(W, codebook, indices_fn, processing, upd_step)

class WeightTVQ_QUIPsharp(WeightTVQ):
  def __init__(self, W, upd_step):
    processing = incoherence_processing
    codebook = quip_sharp_E8_codebook
    indices_fn = fast_E8_rounding
    super().__init__(W, codebook, indices_fn, processing, upd_step)
    
\end{lstlisting}
\end{listing}

\section{Index updates computational costs}\label{subsec: Index updates computational costs}

In TVQ, index matrices are computed according to Equation~\eqref{eq:I_from_W} defined in Section~\ref{subsec: Trainable Vector Quantization}. It requires searching for the index matrix that minimizes the distance between the trainable latent weight and the dequantized weight. The search is performed in the index matrix search space $\displaystyle{\mathcal{I}}$; therefore searching algorithm varies for different VQ schemes. Table~\ref{table: Training Efficiency 2} summarizes the computational resources needed for different search algorithms.

\section{Training acceleration details}\label{subsec: Training acceleration details}
In addition to accelerating index matrix updates, we employ standard techniques to optimize overall training efficiency. For block-wise training in Stage-1 of TVQ, we parallelize index matrices updates inside a Transformer block across multiple processing units. Additionally, we cache intermediate activations $\displaystyle{h_k}$ and $\displaystyle{h_k^q}$ in Random Access Memory to reduce the number of forward passes. We also leverage popular tools such as the JIT-compiling PyTorch API, mixed-precision matrix multiplication, activation checkpointing, and batch accumulation.

\end{document}